%% file: main.tex
\documentclass{article}

\usepackage{xparse}
\usepackage{multirow}
\usepackage{dblfloatfix}
\usepackage{setspace}
\usepackage{amsmath}
\usepackage{amssymb}
\usepackage{amsthm}
\usepackage{caption}
\usepackage{subcaption}
\usepackage{snapshot}

\newsavebox{\fminipagebox}
\NewDocumentEnvironment{fminipage}{m O{\fboxsep}}
 {\par\kern#2\noindent\begin{lrbox}{\fminipagebox}
  \begin{minipage}{#1}\ignorespaces}
 {\end{minipage}\end{lrbox}%
  \makebox[#1]{%
    \kern\dimexpr-\fboxsep-\fboxrule\relax
    \fbox{\usebox{\fminipagebox}}%
    \kern\dimexpr-\fboxsep-\fboxrule\relax
  }\par\kern#2
 }

\usepackage{times}

\usepackage[pdftex]{graphicx}
\usepackage{makecell}

\usepackage{natbib}

\usepackage{algorithm}
\usepackage{algorithmic}

\usepackage{hyperref}

\usepackage[accepted]{icml2017}

\usepackage[multiple]{footmisc}
\usepackage[percent]{overpic}

\input{tools}

\input{macros}

\icmltitlerunning{Adversarial Transformation Networks}

\title{%
  \icmltitle{%
    Adversarial Transformation Networks:\\
    Learning to Generate Adversarial Examples
  }
  \vskip-0.5cm\relax
}

\author{%
  Shumeet Baluja and Ian Fischer\\
  Google Research\\
  Mountain View, CA.
}

\date{}

\begin{document}

\pagenumbering{gobble}
\maketitle
\vskip-0.5cm\relax

\begin{abstract}

  Multiple different approaches of generating adversarial
  examples have been proposed to attack deep neural networks.
  These approaches involve either directly computing gradients with respect to the
  image pixels, or directly solving an optimization on the image pixels.
  In this work, we present a fundamentally new method for generating adversarial examples
  that is fast to execute and provides exceptional diversity of output.
  We efficiently train feed-forward neural networks in a self-supervised manner
  to generate adversarial examples against a target network or set of networks.
  We call such a network an Adversarial Transformation Network (ATN).
  ATNs are trained to generate adversarial examples that minimally modify the
  classifier's outputs given the original input, while constraining the new
  classification to match an adversarial target class.
  We present methods to train ATNs and analyze their effectiveness
  targeting a variety of MNIST classifiers as well as the latest
  state-of-the-art ImageNet classifier Inception ResNet v2.
\end{abstract}

\input{intro}

\input{arch}

\input{mnist}

\input{imagenet}

\section{Conclusions and Future Work}

Current methods for generating adversarial samples involve
a gradient descent procedure on individual input examples.
We have presented a fundamentally different approach to finding examples by
training neural networks to convert inputs into adversarial examples.
Our method is efficient to train, fast to execute, and produces
remarkably diverse, successful adversarial examples.

Future work should explore the possibility of using ATNs in adversarial training.
A successful ATN-based system may pave the way towards models with better
generalization and robustness.

\citet{metzen2017} recently showed that it is possible to detect when
an input is adversarial, for current types of adversaries.
It may be possible to train such detectors on ATN output.
If so, using that signal as an additional loss for the ATN may improve the
outputs.
Similarly, exploring the use of a GAN discriminator during training may
improve the realism of the ATN outputs.
It would be interesting to explore the impact of ATNs on generative models,
rather than just classifiers, similar to work in~\citet{kos2017adversarial}.
Finally, it may also be possible to train ATNs in a black-box manner, similar
to recent work in~\citet{tramer2016stealing,baluja2015virtues}, or using
REINFORCE~\citep{williams1992simple} to compute gradients for the ATN using
the target network simply as a reward signal.

\begingroup
\setstretch{1.0}

\small

\bibliography{icml2017}
\bibliographystyle{icml2017}

\endgroup

\end{document}

%% file: tools.tex
\newcommand{\eat}[1]{}

%% file: macros.tex
\usepackage{amsmath}
\usepackage{amssymb}
\usepackage{amsthm}

\newcommand{\be}{\begin{equation}}
\newcommand{\ee}{\end{equation}}
\newcommand{\bea}{\begin{eqnarray}}
\newcommand{\eea}{\end{eqnarray}}
\newcommand{\beaa}{\begin{eqnarray*}}
\newcommand{\eeaa}{\end{eqnarray*}}
\newcommand{\ba}{\begin{align*}}
\newcommand{\ea}{\end{align*}}

\DeclareMathOperator*{\argmax}{argmax}
\DeclareMathOperator*{\argmin}{argmin}
\DeclareMathOperator*{\onehot}{onehot}

\newcommand{\myvec}[1]{\mathbf{#1}}
\newcommand{\myvecsym}[1]{\boldsymbol{#1}}

\newcommand{\vtheta}{\myvecsym{\theta}}

\newcommand{\vp}{\myvec{p}}

\newcommand{\vx}{\myvec{x}}

\newcommand{\vxp}{\vx\myvec{'}}
\newcommand{\vxstar}{\vx^\myvec{\ast}}

\newcommand{\vy}{\myvec{y}}
\newcommand{\vyp}{\vy\myvec{'}}

\newcommand{\mymathcal}[1]{\mathcal{#1}}
\newcommand{\calG}{\mymathcal{G}}

\newcommand{\calX}{\mymathcal{X}}
\newcommand{\calY}{\mymathcal{Y}}

\newcommand{\hmm}[1]{\ifnum\ifhmode\spacefactor\else2000\fi>1000 \uppercase{#1}\else#1\fi}

\newcommand{\LX}{L_{\calX}}
\newcommand{\LY}{L_{\calY}}

\newcommand{\na}[1]{\textit{IR2-\ifnum0=#1Base-Deconv\else \ifnum1=#1Conv-FC\else \ifnum2=#1Conv-Deconv\else \ifnum3=#1Resize-Conv\else ERROR\fi\fi\fi\fi}}

%% file: intro.tex
\section{Introduction and Background}
\label{intro}

With the resurgence of deep neural networks for many real-world
classification tasks, there is an increased interest in methods to
generate training data, as well as to find weaknesses in trained models.
An effective strategy to achieve both goals is to create
\emph{adversarial examples} that trained models will misclassify.
Adversarial examples are small perturbations of the inputs that are carefully crafted to fool the network into producing incorrect outputs.
These small perturbations can be used both offensively, to fool models
into giving the ``wrong'' answer, and defensively, by providing training
data at weak points in the model.
Seminal work by~\citet{szegedy2013intriguing} and
~\citet{goodfellow2014explaining}, as well as much recent work, has
shown that adversarial examples are abundant, and that there are many
ways to discover them.

Given a classifier $f(\vx): \vx \in \calX \rightarrow y \in \calY$ and original inputs $\vx \in \calX$, the problem of generating \textit{untargeted} adversarial examples can be expressed as the optimization: $\argmin_{\vxstar} L(\vx,\vxstar)\ s.t.\ f(\vxstar) \ne f(\vx)$, where $L(\cdot)$ is a distance metric between examples from the input space (e.g., the $L_2$ norm).
Similarly, generating a \textit{targeted} adversarial attack on a classifier can be expressed as $\argmin_{\vxstar} L(\vx,\vxstar)\ s.t.\ f(\vxstar) = y_t$, where $y_t \in \calY$ is some target label chosen by the attacker.\footnote{
  Another axis to compare when considering adversarial attacks is whether the adversary has access to the internals of the target model.
  Attacks without internal access are possible by transferring successful attacks on one model to another model, as in \citet{szegedy2013intriguing,papernot2016transferability}, and others.
  A more challenging class of blackbox attacks involves having no access to any relevant model, and only getting online access to the target model's output, as explored in \citet{papernot2016practical,baluja2015virtues,tramer2016stealing}.
  See \citet{papernot2015limitations} for a detailed discussion of threat models.
}

Until now, these optimization problems have been solved using three broad approaches:
(1) By directly using optimizers like L-BFGS or Adam \citep{kingma2014adam}, as proposed in \citet{szegedy2013intriguing} and \citet{carlini2016towards}.
Such optimizer-based approaches tend to be much slower and more powerful than the other approaches.  (2) By approximation with single-step gradient-based techniques like fast gradient sign \citep{goodfellow2014explaining} or fast least likely class \citep{kurakin2016adv}.
These approaches are fast, requiring only a single forward and backward pass through the target classifier to compute the perturbation. (3) By approximation with iterative variants of gradient-based techniques \citep{kurakin2016adv, moosavi2016universal, moosavi2016deepfool}.
These approaches use multiple forward and backward passes through the target network to more carefully move an input towards an adversarial classification.

%% file: arch.tex
\section{Adversarial Transformation Networks}
\label{arch}

In this work, we propose Adversarial Transformation Networks (\textit{ATN}s).
An ATN is a neural network that transforms an input into an adversarial example against a target network or set of networks.
ATNs may be untargeted or targeted, and trained in a black-box\footnote{
 E.g., using ~\citet{williams1992simple} to generate training gradients for the ATN based on a reward signal computed on the result of sending the generated adversarial examples to the target network.
}
or white-box manner.
In this work, we will focus on targeted, white-box ATNs.

Formally, an ATN can be defined as a neural network:
\begin{equation}
\label{eqn:atn}
g_{f,\vtheta}(\vx): \vx \in \calX \rightarrow \vxp
\end{equation}
where $\vtheta$ is the parameter vector of $g$, $f$ is the target network which outputs a
probability distribution across class labels, and $\vxp \sim \vx$, but $\argmax f(\vx) \ne \argmax f(\vxp)$.

\paragraph{Training.}
To find $g_{f,\vtheta}$, we solve the following optimization:
\begin{equation}
\label{eqn:atn-train}
\argmin_{\vtheta} \sum_{\vx_i \in \calX} \beta L_{\calX}(g_{f,\vtheta}(\vx_i), \vx_i) + L_{\calY}(f(g_{f,\vtheta}(\vx_i)), f(\vx_i))
\end{equation}
where $L_{\calX}$ is a loss function in the input space (e.g., $L_2$ loss or a perceptual similarity loss like \citet{johnson2016perceptual}),
$L_{\calY}$ is a specially-formed loss on the output space of $f$ (described below) to avoid learning the identity function, and $\beta$ is a weight to balance the two loss functions.
We will omit $\vtheta$ from $g_f$ when there is no ambiguity.

\vskip -0.1in
\paragraph{Inference.}

At inference time, $g_f$ can be run on any input $\vx$
without requiring further access to $f$ or more gradient computations.
This means that after being trained, $g_f$ can generate adversarial examples against the
target network $f$ even faster than the single-step gradient-based
approaches, such as fast gradient sign, so long as $||g_f|| \lessapprox ||f||$.

\vskip -0.1in
\paragraph{Loss Functions.}
The input-space loss function, $L_{\calX}$, would ideally correspond
closely to human perception.
However, for simplicity, $L_2$ is sufficient.
$L_{\calY}$ determines whether or not the ATN is targeted; the \emph{target} refers to the class for which the
adversary will cause the classifier to output the maximum value.
In this work, we focus on the more challenging case of creating targeted ATNs, which can be defined similarly
to Equation~\ref{eqn:atn}:
\begin{equation}
\label{eqn:atn-t}
g_{f,t}(\vx): \vx \in \calX \rightarrow \vxp
\end{equation}
where $t$ is the target class, so that $\argmax f(\vxp)=t$.  This allows us to
target the exact class the classifier should mistakenly believe the
input is.

In this work, we define $L_{\calY,t}(\vyp,\vy)=L_2(\vyp, r(\vy,t))$, where $\vy=f(\vx)$, $\vyp=f(g_f(\vx))$, and $r(\cdot)$ is a reranking function that modifies $\vy$ such that $y_k < y_t, \forall~k \neq t$.

Note that training labels for the target network are not required at any point in this process.
All that is required is the target network's outputs $\vy$ and $\vyp$.
It is therefore possible to train ATNs in a self-supervised manner, where they use unlabeled data as the input and make $\argmax~f(g_{f,t}(\vx))=t$.

\paragraph{Reranking function.}
There are a variety of options for the reranking function.
The simplest is to set $r(\vy,t)=\onehot(t)$, but other formulations can make better use of the signal already present in $\vy$ to encourage better reconstructions.
In this work, we look at reranking functions that attempt to keep $r(\vy,t) \sim \vy$.
In particular, we use $r(\cdot)$ that maintains the rank order of all but the targeted class in order to minimize distortions when computing $\vxp=g_{f,t}(\vx)$.

The specific $r(\cdot)$ used in our experiments has the following form:
\begin{equation}
\label{eqn:atn-rerank}
r_{\alpha}(\vy,t) = norm\left(
\left\{
\begin{aligned}
    \alpha * \max \vy && \text{if } k = t \\
    y_k               && \text{otherwise}
\end{aligned}
\right\}_{k \in \vy}
\right)
\end{equation}
$\alpha>1$ is an additional parameter specifying how much larger $y_t$ should be than the current max classification.
$norm(\cdot)$ is a normalization function that rescales its input to be a valid probability distribution.

\subsection{Adversarial Example Generation}
\label{generation}

There are two approaches to generating adversarial examples with an ATN.
The ATN can be trained to generate just the perturbation to $\vx$, or it can be trained to generate an \textit{adversarial autoencoding} of $\vx$.

\begin{itemize}
\item \textbf{Perturbation ATN (\textit{P-ATN}):}
To just generate a perturbation, it is sufficient to structure the ATN as a variation on the residual block \citep{he2015resnet}:
$g_f(\vx)=\tanh (\vx+\calG(\vx))$, where $\calG(\cdot)$ represents the core function of $g_f$.
With small initial weight vectors, this structure makes it easy for the network to learn to generate small, but effective, perturbations.

\item \textbf{Adversarial Autoencoding (\textit{AAE}):}
AAE ATNs are similar to standard autoencoders, in that they
attempt to accurately reconstruct the original input, subject to
regularization, such as weight decay or an added noise signal.
For AAE ATNs, the regularizer is $L_{\calY}$.
This imposes an additional requirement on the AAE to add some perturbation $\vp$ to
$\vx$ such that $r(f(\vxp))=\vyp$.

\end{itemize}

For both ATN approaches, in order to enforce that $\vxp$ is a
plausible member of $\calX$, the ATN should only generate values in
the valid input range of $f$.  For images, it suffices to set the
activation function of the last layer to be the $tanh$ function; this
constrains each output channel to $[-1,1]$.

\subsection{Related Network Architectures}

This training objective resembles standard
Generative Adversarial Network training~\cite{goodfellow2014generative} in that
the goal is to find weaknesses in the classifier.
It is interesting to note the similarity to  work outside the adversarial
training paradigm ---  the recent use of feed-forward neural networks for artistic style transfer in images \cite{gatys2015}\cite{ulyanovLVL16}.
\citet{gatys2015} originally proposed  a gradient descent
procedure based on ``back-driving networks''~\cite{linden1989inversion} to modify the inputs of a fully-trained network to
find a set of inputs that maximize a desired set of outputs and
hidden unit activations.
Unlike standard network training in
which the gradients are used to modify the weights of the network,
here, the network weights are frozen and the input itself is changed.
In subsequent work, \citet{ulyanovLVL16} created a method to approximate the results of the gradient descent procedure through the use of an off-line trained neural network.
\citet{ulyanovLVL16} removed the need for a gradient descent procedure to operate on every source image to which a new artistic style was to be applied, and replaced it with a single forward pass through a separate network.  Analagously, we do the same for generating adverarial examples: a separately trained network approximates the usual gradient descent procedure done on the target network to find adversarial examples.

%% file: mnist.tex
\begin{table}
\center
\caption{Baseline Accuracy of Five MNIST Classifiers}
\label{table:mnistTrainPerf}
\footnotesize

\begin{tabular}{lr}
\hline
  Architecture & Acc.  \\
\hline

\makecell[l] {Classifier-Primary (Classifier$_p$) \\
~~(5x5 Conv)$\rightarrow$ (5x5 Conv)$\rightarrow$ FC$\rightarrow$ FC } & 98.6\% \\
\makecell[l]{Classifier-Alternate-0 (Classifier$_{a0}$) \\
~~(5x5 Conv)$\rightarrow$ (5x5 Conv)$\rightarrow$ FC $\rightarrow$ FC }& 98.5\% \\

\makecell[l] {Classifier-Alternate-1 (Classifier$_{a1}$) \\
~~(4x4 Conv)$\rightarrow$ (4x4 Conv)$\rightarrow$ (4x4 Conv)$\rightarrow$ FC $\rightarrow$ FC }& 98.9\% \\

\makecell[l] {Classifier-Alternate-2 (Classifier$_{a2}$) \\
~~(3x3 Conv)$\rightarrow$ (3x3 Conv)$\rightarrow$ (3x3 Conv) $\rightarrow$ FC $\rightarrow$ FC }& 99.1\% \\

\makecell[l] {Classifier-Alternate-3 (Classifier$_{a3}$) \\
~~(3x3 Conv) $\rightarrow$ FC $\rightarrow$ FC $\rightarrow$ FC }& 98.5\% \\

\hline
\end{tabular}

\end{table}

\section{MNIST Experiments}
\label{mnist}

To begin our empirical exploration, we train five networks on the standard MNIST
digit classification task~\cite{lecun1998mnist}.
The networks are trained and tested on the same data; they vary only in the weight
initialization and architecture, as shown in Table~\ref{table:mnistTrainPerf}.
Each network has a mix of convolution (Conv) and Fully Connected (FC) layers.
The input to the networks is a 28x28 grayscale image and the output is 10 logit units.
Classifier$_p$ and Classifier$_{a0}$ use the same architecture, and only differ
in the initialization of the weights.
We will primarily use Classifier$_p$ for the experiments in this section.
The other networks will be used later to analyze the generalization capabilities
of the adversaries.
Table~\ref{table:mnistTrainPerf} shows that all of the networks perform well
on the digit recognition task.\footnote{%
  It is easy to get better performance than this on MNIST, but for these
  experiments, it was more important to have a variety of architectures that
  achieved similar accuracy, than to have state-of-the-art performance.
}

\begin{figure}
  \begin{center}
    \begin{minipage}{0.8in}
    \includegraphics[width=0.8in]{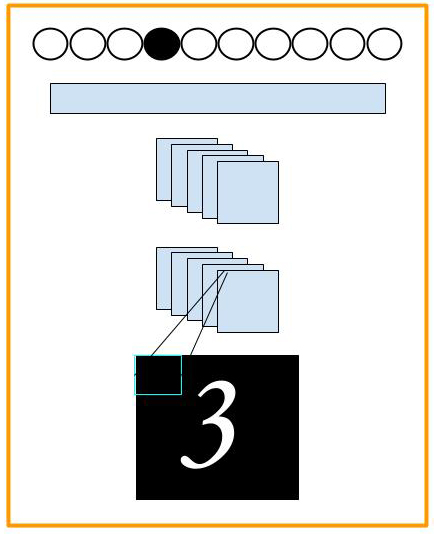}
    \end{minipage}
    ~~~~~~
    \begin{minipage}{1.1in}
      \includegraphics[width=1.05in]{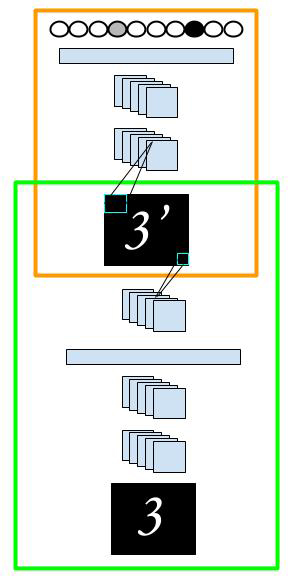}\\
    \end{minipage}
    \caption{%
      (Left) A simple classification network which takes input image $\vx$.
      (Right) With the same input, $\vx$, the ATN emits $\vxp$,
      which is fed into the classification network.
      In the example shown, the input digit is classified correctly as a 3
      (on the left),
      ATN$_7$ takes $\vx$ as input and generates a modified image ($3'$)
      such that the classifier outputs a 7 as the highest activation
      and the previous highest classification, 3, as the second highest
      activation (on the right).
      \label{fig:adversaryArch}
    }
  \end{center}
\end{figure}

We attempt to create an Adversarial Autoencoding ATN that can target a
specific class given any input image.
The ATN is trained against a particular classifier as illustrated in
Figure~\ref{fig:adversaryArch}.
The ATN takes the original input image, $\vx$, as input, and outputs a new
image, $\vxp$, that the target classifier should erroneously classify as $t$.
We also add the constraint that the ATN should maintain the ordering
of all the other classes as initially output by the classifier.  We
train ten ATNs against Classifier$_p$ -- one for each target digit,
$t$.

\begin{table*}
\center

\caption{%
  Average success of ATN$_{0-9}$ at transforming an image such that it is misclassified by
  Classifier$_p$.
  As $\beta$ is reduced, the ability to fool Classifier$_p$ increases.    How to read the table:
  Top row of cell: percentage of times Classifier$_p$ labeled $\vxp$ as $t$.
  Middle row of cell: percentage of times Classifier$_p$ labeled $\vxp$ as $t$ and kept the
  original classification ($\argmax \vy$) in second place.
  Bottom row of cell: percentage of all $\vxp$ that kept the original classification in second place.}
\label{table:adversaryPerf}

\vskip 0.15in
\begingroup
\setlength{\tabcolsep}{6pt} % Default value: 6pt
\renewcommand*{\arraystretch}{1} % Default value: 1

\begin{tabular}{l|cccc}

&  \multicolumn{3}{c}{$\beta$:} \\
&  0.010 & 0.005 & 0.001 \\

\hline
\makecell[l]{\textbf{ATN$_a$} \\ FC $\rightarrow$ FC$\rightarrow$28x28 Image\\}  &

\makecell{\textbf{69.1\%}\\ \small 91.7\% \\ \small 63.5\%}& \makecell{\textbf{84.1\%}\\ \small 93.4\% \\ \small 78.6\%}& \makecell{\textbf{95.9\%}\\ \small 95.3\% \\ \small 91.4\%} \\

\hline

\makecell[l]{\textbf{ATN$_b$} \\ (3x3 Conv)$\rightarrow$ (3x3 Conv)  $\rightarrow$\\ (3x3 Conv)
 $\rightarrow$ FC $\rightarrow$ 28x28 Image~\\} &
\makecell{\textbf{61.8\%}\\ \small 93.8\% \\ \small 58.7\%}& \makecell{\textbf{77.7\%}\\ \small 95.8\% \\ \small 74.5\%}& \makecell{\textbf{89.2\%}\\ \small 97.4\% \\ \small 86.9\%} \\

\hline
 \makecell[l]{\textbf{ATN$_c$} \\ (3x3 Conv)$\rightarrow$ (3x3 Conv)$\rightarrow$(3x3 Conv)
  \\ $\rightarrow$ Deconv: 7x7  $\rightarrow$ Deconv: 14x14 $\rightarrow$  28x28 Image} &
\makecell{\textbf{66.6\%}\\ \small 95.5\% \\ \small 64.0\%}& \makecell{\textbf{82.5\%}\\ \small 96.6\% \\ \small 79.7\%}& \makecell{\textbf{91.4\%}\\ \small 97.5\% \\ \small 89.1\%} \\

\hline
\end{tabular}
\endgroup

\vskip 0.3in
\end{table*}

An example is provided to make this concrete. If a classifier is given
an image, $\vx_3$, of the digit 3, a successful ordering of the outputs
(from largest to smallest) may be as follows: Classifier$_p(\vx_3)\rightarrow[3,8,5,0,4,1,9,7,6,2]$.
If ATN$_7$ is applied to $\vx_3$, when the resulting image, $\vxp_3$, is fed into the same classifier,
the following ordering of outputs is desired (note that the 7 has
moved to the highest output):
Classifier$_p($ATN$_7(\vx_3))\rightarrow[7,3,8,5,0,4,1,9,6,2]$.

Training for a single ATN$_t$ proceeds as follows.
The weights of Classifier$_p$ are frozen and never change during ATN training.
Every training image, $\vx$, is passed through Classifier$_p$ to obtain output $\vy$.
As described in Equation~\ref{eqn:atn-rerank}, we then compute $r_{\alpha}(\vy,t)$
by copying $\vy$ to a new value, $\vyp$, setting $y_t'=\alpha * \max(\vy)$,
and then renormalizing $\vyp$ to be a valid probability distribution.
This sets the target class, $t$, to have the highest value in $\vyp$
while maintaining the relative order of the other original classifications.
In the MNIST experiments, we empirically set $\alpha=1.5$.

Given $\vyp$, we can now train ATN$_t$ to generate $\vxp$ by minimizing
$\beta * \LX=\beta * L_2(\vx,\vxp)$ and $\LY=L_2(\vy,\vyp)$ using Equation~\ref{eqn:atn-train}.
Though the weights of Classifier$_p$ are frozen, error derivatives are still passed through them to train the ATN.
We explore several values of $\beta$ to balance the two loss functions.
The results are shown in Table~\ref{table:adversaryPerf}.

\paragraph{Experiments.}
We tried three ATN architectures for the AAE task, and each
was trained with three values of $\beta$ against all ten targets, $t$.
The full $3\times3$ set of experiments are shown in Table~\ref{table:adversaryPerf}.
The accuracies shown are the ability of ATN$_t$ to transform an input image
$\vx$ into $\vxp$ such that Classifier$_p$ mistakenly classifies $\vxp$ as
$t$.\footnote{%
  Images that were originally classified as $t$ were not
  counted in the test as no transformation on them was
  required.
} 
Each measurement in Table~\ref{table:adversaryPerf} is the average of the 10
networks, ATN$_{0-9}$.

\begin{figure}
  \begin{center}
    \includegraphics[width=3.5in]{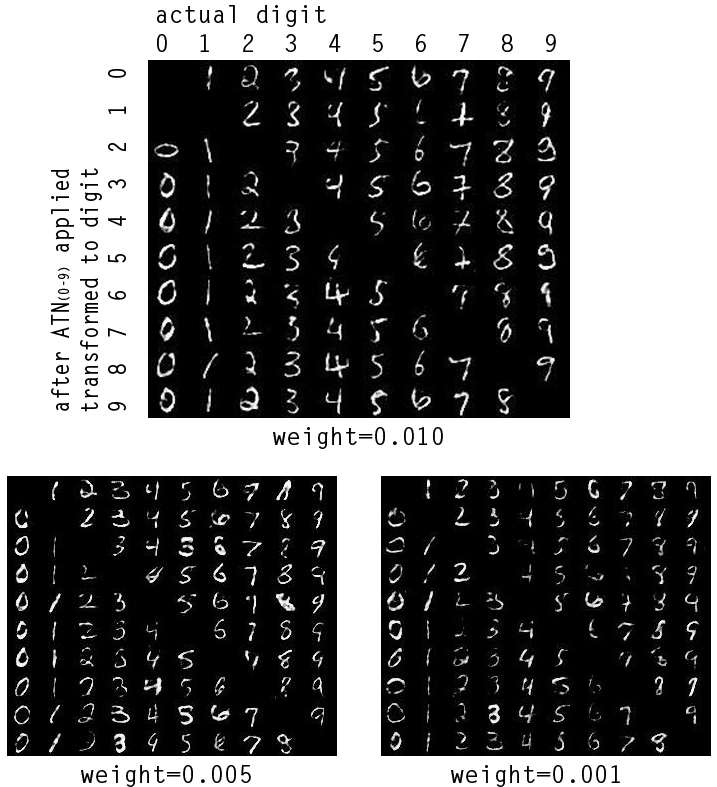}\\
    \caption{%
      Successful adversarial examples from ATN$_t$ against Classifier$_p$.
      Top is with the highest $\beta=0.010$.
      Bottom two are with $\beta=0.005~\&~0.001$, respectively.
      Note that as $\beta$ is decreased, the fidelity to the underlying digit decreases.
      The column in each block corresponds to the correct classification
      of the image.
      The row corresponds to the adversarial classification, $t$.
      \label{fig:exampleTransforms}
    }
  \end{center}

  \vspace{\floatsep}
  \vspace{-0.1in}
\end{figure}

\paragraph{Results.}
In Figure~\ref{fig:exampleTransforms}(top), each row represents the
transformation that ATN$_t$ makes to digits that were initially
correctly classified as 0-9 (columns).
For example, in the top row, the digits 1-9 are now all classified as 0.
In all cases, their second highest classification is the original correct
classification (0-9).

\begin{figure*}
  \vskip 0.1in
  
  \begin{center}
    \includegraphics[width=5.8in]{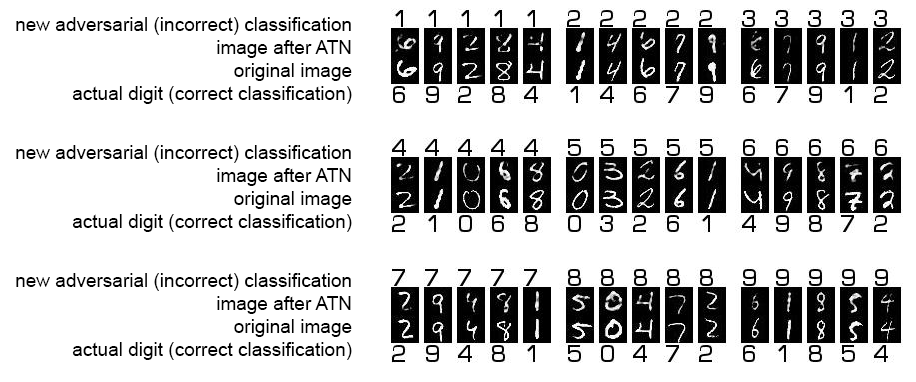}
    \caption{%
      Typical transformations made to MNIST digits against Classifier$_p$.
      Black digits on the white background are output classifications from Classifier$_p$.
      The bottom classification is the original (correct) classification.
      The top classification is the result of classifying the adversarial example.
      White digits on black backgrounds are the MNIST digits and their transformations to adversarial examples.
      The bottom MNIST digits are unmodified, and the top are adversarial.
      In all of these images, the adversarial example is classified as $t=\argmax \vyp$ while maintaining
      the second highest output in $\vyp$ as the original classification, $\argmax \vy$.
      \label{fig:beforeAfter}
    }
  \end{center}
  \vskip 0.3in
\end{figure*}

The reconstructions shown in Figure~\ref{fig:exampleTransforms}(top)
have the largest $\beta$; smaller $\beta$ values are shown in the
bottom row.
The fidelity to the underlying digit diminishes as $\beta$ is reduced.
However, by loosening the constraints to stay similar to the original input, the
number of trials in which the transformer network is able to successfully
``fool'' the classification network increases dramatically, as seen in
Table~\ref{table:adversaryPerf}.
Interestingly, with 
$\beta = 0.010$, in
Figure~\ref{fig:exampleTransforms}(second row), where
there should be a `0' that is transformed into a `1', no digit
appears.
With this high $\beta$, no example was found
that could be transformed to successfully fool Classifier$_p$.
With the two smaller $\beta$ values, this anomaly does not occur.

In Figure~\ref{fig:beforeAfter}, we provide a closer look at examples of
$\vx$ and $\vxp$ for ATN$_c$ with $\beta=0.005$.
A few points should be noted:
\begin{itemize}
  \item{%
    The transformations maintain the large, empty regions of the
    image.  Unlike many previous studies in attacking classifiers, the
    addition of salt-and-pepper type noise did not
    appear~\cite{nguyen2014,moosavi2016deepfool}.
  }

  \item{%
    In the majority of the generated examples, the shape of the
    digit does not dramatically change.  This is the desired behavior:
    by training the networks to maintain the order beyond the
    top-output, only minimal changes should be made to the image.
    The changes that are often introduced are patches where the light
    strokes have become darker.
  }

  \item{%
    Vertical-linear components of the original images are
    emphasized in several digits; it is especially noticeable in the
    digits transformed to 1.  With other digits (e.g., 8), it is
    more difficult to find a consistent pattern of what is being
    (de)emphasized to cause the classification network to be fooled.
  }
\end{itemize}

A novel aspect of ATNs is that though they cause the target classifier
to output an erroneous top-class, they are also trained to ensure that
the transformation preserves the existing output ordering of the
target-classifier (other than the top-class).  For the examples that
were successfully transformed, Table~\ref{table:rankDiff} gives the
average rank-difference of the outputs with the pre-and-post
transformed images (excluding the intentional targeted misclassification).

\begin{table}
  \begin{center}
  \begin{small}
    \caption{%
      Rank Difference in Secondary Outputs, Pre/Post Transformation.
      Top-5 (Top-9).
      \label{table:rankDiff}
    }

    \begin{tabular}{l|ccc}
      & \multicolumn{3}{c}{$\beta$:} \\
      &  0.010  & 0.005 & 0.001 \\
      \hline
      ATN$_a$ & 0.93 (0.99) & 0.98 (1.04) & 1.04 (1.13)  \\
      \hline
      ATN$_b$ & 0.81 (0.87) & 0.83 (0.89) & 0.86 (0.93)  \\
      \hline
      ATN$_c$ & 0.79 (0.85) & 0.83 (0.90) & 0.89 (0.97)  \\
      \hline
    \end{tabular}

  \end{small}
  \end{center}
  \vspace {0.1in}
\end{table}

\begin{table*}
  %  \vskip -.1in
  \begin{center}
    \captionof{table}{%
      ATN$_b$ with $\beta=0.005$ trained to defeat Classifier$_p$.
      Tested on 5 classifiers, without further training, to measure transfer.
      1st place is the percentage of times $t$ was the top classification.
      2nd place measures how many times the original top class ($\argmax \vy$) was correctly placed into 2nd place,
      conditioned on the 1st place being correct (Conditional) or unconditioned on 1st place (Unconditional).
      \label{table:adversaryVSalternates}
    }
    \vspace{0.2in}
    \begin{tabular}{r|ccccc}

      & Classifier$_p$* & Classifier$_{a0}$ & Classifier$_{a1}$ & Classifier$_{a2}$ & Classifier$_{a3}$ \\
      \hline
      1st Place Correct &  82.5\% & 15.7\% & 16.1\% & 7.7\% & 28.9\%  \\
      %\hline
      \small 2nd Place Correct (Conditional) &  \small 96.6\% & \small
      84.7\% & \small 89.3\% & \small 85.0\% & \small 81.8\%  \\

      %\hline
      \small 2nd Place Correct (Unconditional)  & \small 79.7\%
      & \small 15.6\% & \small 16.1\% & \small 8.4\% &
      \small 26.2\%  \\

      \hline
    \end{tabular}

  \end{center}
\vspace{0.2in}
\end{table*}

\section{A Deeper Look into ATNs}
\label{deeperLook}

This section explores three extensions to the basic ATNs:
increasing the number of networks the ATNs can attack,
using hidden state from the target network,
and using ATNs in serial and parallel.

\subsection{Adversarial Transfer to Other Networks}
\label{transfer}

So far, we have examined ATNs in the context of attacking a single
classifier.
Can ATNs create adversarial examples that generalize to other classifiers?
Much research has studied adversarial transfer for traditional adversaries,
including the recent work of \citet{moosavi2016universal,liu2016delving}.

\paragraph{Targeting multiple networks.}
To test transfer, we take the adversarial examples from the previously trained ATNs
and test them against Classifier$_{a0,a1,a2,a3}$ (described in
Table~\ref{table:mnistTrainPerf}).

The results in Table~\ref{table:adversaryVSalternates} clearly show that the
transformations made by the ATN \emph{are not} general; they are tied
to the network it is trained to attack.
Even Classifier$_{a0}$, which has the same architecture as Classifier$_p$,
is not more susceptible to the attacks than those with different architectures.
Looking at the second place correctness scores (in the same
Table~\ref{table:adversaryVSalternates}), it may, at first, seem
counter-intuitive that the conditional probability of a correct
second-place classification remains high despite a low first-place
classification.
The reason for this is that in the few cases in which the
ATN was able to successfully change the classifier's top choice, the
second choice (the real classification) remained a close second
(i.e., the image was not transformed in a large manner), thereby
maintaining the high performance in the conditional second rank
measurement.

\begin{figure}
  \begin{center}
    \includegraphics[width=3.2in]{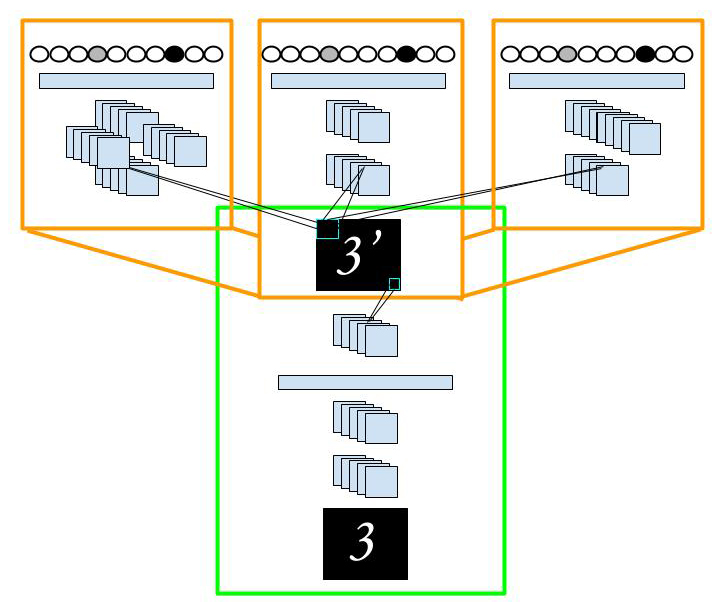}\\
    \caption{%
      The ATN now has to fool three networks (of various architectures),
      while also minimizing $\LX$, the reconstruction error.
      \label{fig:transferanceArch}
    }
  \end{center}
\end{figure}

\paragraph{Training against multiple networks.}
Is it possible to create a network that will be able to create a
single transform that can attack \emph{multiple} networks?
Will such an ATN generalize better to unseen networks?
To test this, we created an ATN that receives training signals from multiple networks, as shown
in Figure~\ref{fig:transferanceArch}. As with the earlier training,
the $\LX$ reconstruction error remains.

\begin{table*}
  % \vskip 0.15in
  \begin{center}
    \captionof{table}{%
      ATN$_b$ retrained with 3 networks (marked with *).
      \label{table:adversaryRetrain3}
    }
    \begin{tabular}{c|r|cccccc}

      \hline

      $\beta$ &  & Classifier$_p$* & Classifier$_{a0}$ & Classifier$_{a1}$* & Classifier$_{a2}$* & Classifier$_{a3}$ \\
      \hline
      \multirow{3}{1cm}{0.010} & 1st Place Correct &  89.9\% & 37.9\% &
      83.9\% & 78.7\% & 70.2\% & \\
      &\small 2nd Place Correct (Conditional) &\small 96.1\% &\small 88.1\% &\small 96.1\% &\small 95.2\% &\small 79.1\% &\small\\
      &\small 2nd Place Correct (Unconditional)  &\small 86.4\% &\small 34.4\% &\small 80.7\% &\small74.9\% &\small 55.9\% &\small \\

      \hline
      \hline
      \multirow{3}{1cm}{0.005} &1st Place Correct &  93.6\% & 34.7\% & 88.1\% & 82.7\% & 64.1\% & \\
      &\small 2nd Place Correct (Conditional) &\small 96.8\% &\small 88.3\% &\small 96.9\% &\small 96.4\% &\small 73.1\% &\small \\
      &\small 2nd Place Correct (Unconditional)  &\small 90.7\% &\small 31.4\% &\small 85.3\% &\small 79.8\% &\small 47.2\% & \\
      \hline
    \end{tabular}
  \end{center}

\vskip 0.15in

\end{table*}

The new ATN was trained with classification signals from three
networks: Classifier$_p$, and Classifier$_{a1,2}$.
The training proceeds in exactly the same manner as described earlier, except
the ATN attempts to minimize $\LY$ for all three target networks at the same time.
The results are shown in Table~\ref{table:adversaryRetrain3}.
First, examine the columns corresponding to the networks that were used in
the training (marked with an~*).
Note that the success rates of attacking these three classifiers are
consistently high, comparable with those when the ATN was trained with a
single network.
Therefore, it is possible to learn a transformation network that modifies
images such that perturbation defeats multiple networks.

Next, we turn to the remaining two networks to which the adversary was
\emph{not} given access during training.
There is a large increase in success rates over those when the ATN was
trained with a single target network (Table~\ref{table:adversaryVSalternates}).
However, the results do not match those of the networks used in training.
It is possible that training against larger numbers of target networks at
the same time could further increase the transferability of the adversarial
examples.

\begin{figure}
  \begin{center}

    \includegraphics[width=2in, height=1in]{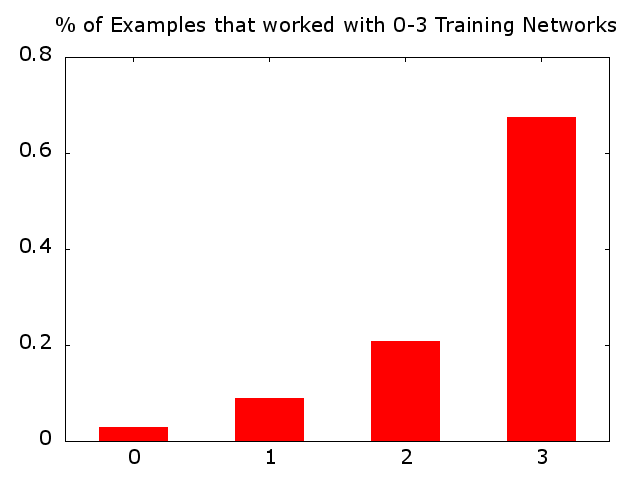}\\
    \includegraphics[width=2in, height=1in]{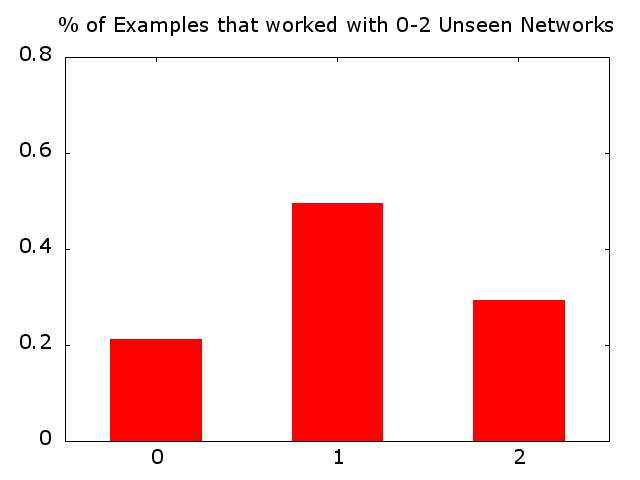}\\

  \caption{%
    Do the same transformed examples work well on all the networks?
    (Top) Percentage of examples that worked on exactly 0-3 training networks.
    (Bottom) Percentage of examples that worked on exactly 0-2 unseen networks.
    Note: these are all measured on independent test set images.
    \label{fig:transformedStats}
  }
  \end{center}
\end{figure}

Finally, we look at the success rates of image transformations.  Do
the same images consistenly fool the networks, or are the failure
cases of the networks different?  As shown in
Figure~\ref{fig:transformedStats}, for the 3 networks the ATN was
trained to defeat, the majority of transformations attacked all three
networks successfully.  For the unseen networks, the results were
mixed; the majority of transformations successfully attacked only a
single network.

\subsection{``Insider'' Information}
\label{insider}

In the experiments thus far, the classifier, $C$, was treated as a
white box. From this box, two pieces of information were needed to
train the ATN.  First, the actual outputs of $C$ were used to create
the new target vector.  Second, the error derivatives from the new
target vector were passed through $C$ and propagated into the ATN.

In this section, we examine the possibility of ``opening'' the
classifier, and accessing more of its internal state.  From $C$, the
actual \emph{hidden unit activations} for each example are used as
additional inputs to the ATN.  Intuitively, because the goal is to
maintain as much similarity as possible to the original image and to
maintain the same order of the non-top-most classifications as the
original image, access to these activations may convey usable signals.

Because of the very large number of hidden units that accompany
convolution layers, in practice, we only use the penultimate
fully-connected layer from $C$.  The results of training the ATNs with
this extra information are shown in
Table~\ref{table:insiderInfoPerformance}.  Interestingly, the most
salient difference does not come from the ability of the ATN to attack
the networks in the first-position.  Rather, when looking at the
conditional-successes of the second-position, the numbers are improved
(compare to Table~\ref{table:adversaryPerf}).  We speculate that this
is because the extra hints provided by the classifier's internal
activations (with the unmodified image) could be used to also ensure
that the second-place classification, after input modification, was
also correctly maintained.

\begin{table}[h]
\centering
    \captionof{table}{%
      Using the internal states of the classifier \emph{as inputs} for the
      Adversary Networks.
      Larger font is the percentage of times the adversarial class was
      classified in the top-space.
      Smaller font is how many times the original top class was correctly
      placed into 2nd place, conditioned on the 1st place being correct
      or not.
      \label{table:insiderInfoPerformance}
    }
    \vskip 0.2in
    \begin{tabular}{l|c|c|c}

      \belowspace
       &  \multicolumn{3}{c}{$\beta$:}  \\
      &  0.010  & 0.005 & 0.001\\
      \hline

      \belowspace
      ATN$_a$ &
      \makecell{68.0\%\\ \scriptsize (94.5\%/64.5\%)}& \makecell{81.4\%\\ \scriptsize (96.0\%/78.1\%)}& \makecell{95.4\%\\ \scriptsize (98.1\%/93.6\%)} \\

      \hline
      \belowspace
      ATN$_b$ &
      \makecell{68.1\%\\ \scriptsize (96.9\%/66.5\%)}& \makecell{78.9\%\\ \scriptsize (98.1\%/77.4\%)}& \makecell{92.4\%\\ \scriptsize (98.9\%/91.4\%)} \\

      \hline
      \belowspace
      ATN$_c$ &
      \makecell{67.9\%\\ \scriptsize (97.6\%/66.4\%)}& \makecell{81.0\%\\ \scriptsize (98.2\%/79.5\%)}& \makecell{93.1\%\\ \scriptsize (99.0\%/92.1\%)} \\

      \hline
    \end{tabular}
    \vskip 0.2in    
\end{table}

\begin{figure*}
  % \vspace {0.5in}
  \begin{center}
  \begin{small}

    \rotatebox{90}{\sc Parallel}~~
    \begin{minipage}{1.7in}
      \includegraphics[width=1.7in]{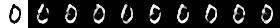}\\
      \includegraphics[width=1.7in]{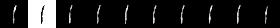}\\
      \includegraphics[width=1.7in]{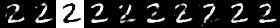}\\
      \includegraphics[width=1.7in]{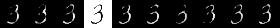}\\
      \includegraphics[width=1.7in]{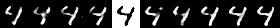}\\
      \includegraphics[width=1.7in]{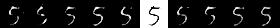}\\
      \includegraphics[width=1.7in]{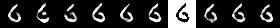}\\
      \includegraphics[width=1.7in]{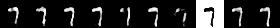}\\
      \includegraphics[width=1.7in]{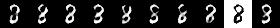}\\
      \includegraphics[width=1.7in]{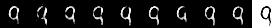}\\
    \end{minipage}
    ~~~~~~
    \begin{minipage}{1.2in}  
      \includegraphics[height=1.5in]{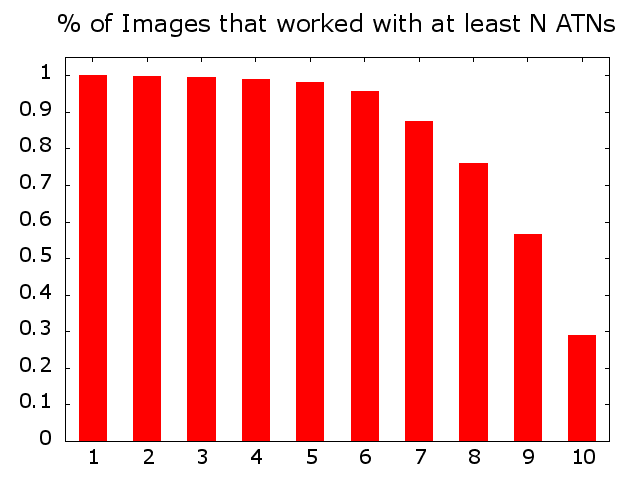}
    \end{minipage}
    ~~~~~~~~~~~~~~~~~~~~~~~~~~~~~~~~~~~~ \it{vs.} ~~~~~~~~
    \rotatebox{90}{\sc Serial}~~
    \begin{minipage}{1.7in}
      \includegraphics[width=1.7in]{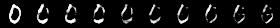}\\
      \includegraphics[width=1.7in]{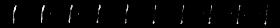}\\
      \includegraphics[width=1.7in]{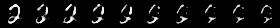}\\
      \includegraphics[width=1.7in]{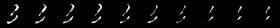}\\
      \includegraphics[width=1.7in]{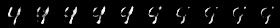}\\
      \includegraphics[width=1.7in]{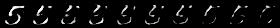}\\
      \includegraphics[width=1.7in]{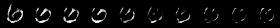}\\
      \includegraphics[width=1.7in]{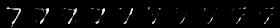}\\
      \includegraphics[width=1.7in]{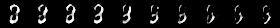}\\
      \includegraphics[width=1.7in]{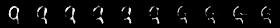}\\
    \end {minipage}

    \vskip -0.1in
    \caption{%
      Parallel and Serial Application of 10 ATNs.
      \textit{Left:}
      Examples of the same original image (shown in white background)
      transformed correctly by all ATNs.
      For example, in the row of 7s, in the first column, the 7 was
      transformed such that the classifier output a 0 as top class, in the
      second column, the classifier output a 1, etc.
      \textit{Middle:}
      Histogram showing the number of images that were transformed
      successfully with at least $N$ ATNs (1-10) when used in parallel.
      \textit{Right:}
      Serial Adversarial Transformation Networks.
      In the first column, ATN$_0$ is applied to the input image.
      In the second column, ATN$_1$ is applied to the
      output of ATN$_0$, etc.
      In each of these examples, all 10 of the ATNs successfully transformed
      the previous image to fool the classifier.
      Note the severe image degradation as the transformation networks are
      applied in sequence.
      \label{fig:parallel}
    }

  \end{small}
  \end{center}

\end{figure*}

\subsection {Serial and Parallel ATNs}

Separate ATNs are created for each digit (0-9).  In this section, we
examine whether the ATNs can be used in parallel (can the same
original image be transformed by each of the ATNs successfully?)  and
in serial (can the same image be transformed by one ATN then that
resulting image be transformed by another, successfully?).

In the first test, we started with 1000 images of digits from the test
set.
Each was passed through all 10 ATNs (ATN$_c$, $\beta=0.005$);
the resulting images were then classified with Classifier$_p$.
For each image, we measured how many ATNs were able to successfully
transform the image (success is defined for ATN$_t$ as causing the
classifier to output $t$ as the top-class).
Out of the 1000 trials, 283 were successfully transformed by all 10
of the ATNs.
Samples results and a histogram of the results are shown in Figure~\ref{fig:parallel}.

A second experiment is constructed in which the 10 ATNs are applied
\emph{serially, one-after-the-other}.
In this scenario, first ATN$_0$ is applied to image $\vx$, yielding $\vxp$.
Then ATN$_1$ is applied to $\vxp$ yielding $\vxp'$ ... to ATN$_9$.
The goal is to see whether the transformations work on previously
transformed images.
The results of chaining the ATNs together in this manner are shown in
Figure~\ref{fig:parallel}(right).
The more transformations that are applied, the larger the image degradation.
As expected, by the ninth transformation (rightmost column in
Figure~\ref{fig:parallel}) the majority of images are severely degraded and
usually not recognizable.
Though we expected the degradation in images, there were two additional,
surprising, findings.
First, in the parallel application of ATNs (the first experiment
described above), out of 1000 images, 283 of them were successfully
transformed by 10 of the ATNs.
In this experiment, 741 images were successfully transformed by 10 ATNs.
The improvement in the number of all-10 successes over applying the ATNs
in parallel occurs because each transformation effectively diminishes
the underlying original image (to remove the real classification from
the top-spot).
Meanwhile, only a few new pixels are added by the ATN to cause the
misclassification as it is also trained to minimize the reconstruction error.
The overarching effect is a fading of the image through chaining ATNs together.

Second, it is interesting to examine what happens to the second-highest
classifications that the networks were also trained to preserve.
Order preservation \emph{did not} occur in this test.
Had the test worked perfectly, then for an input-image, $\vx$ (e.g.,
of the digit 8), after ATN$_0$ was applied, the first and second top
classifications of $\vxp$ should be 0,8, respectively.
Subsequently, after ATN$_1$ is then applied to $\vxp$, the
classifications of $\vxp'$
should be 1,0,8, etc.
The reason this does not hold in practice is that though the networks
were trained to maintain the high classification (8) of the original
digit, $\vx$, they were \emph{not trained to maintain the potentially
small perturbations} that ATN$_0$ made to $\vx$ to achieve a
top-classification of 0.
Therefore, when ATN$_1$ is applied, the changes that ATN$_0$ made may
not survive the transformation.
Nonetheless, if chaining adversaries becomes important, then training
the ATNs with images that have been previously modified by other ATNs
may be a sufficient method to address the difference in training and
testing distributions.
This is left for future work.

%% file: imagenet.tex
\section{ImageNet Experiments}
\label{imagenet}

We explore the effectiveness of ATNs on the ImageNet dataset~\citep{deng2009imagenet},
which consists of 1.2 million natural images categorized into 1 of 1000 classes.
The target classifier, $f$, used in these experiments is a pre-trained state-of-the-art
classifier, Inception ResNet v2 (IR2), that has a top-1 single-crop error rate of 19.9\%
on the 50,000 image validation set, and a top-5 error rate of 4.9\%.
It is described fully in \citet{szegedy2016inception}.

\subsection{Experiment Setup}

We trained AAE ATNs and P-ATNs as described in Section~\ref{arch} to attack IR2.
Training an ATN against IR2 follows the process described in Section~\ref{mnist}.

IR2 takes as input images scaled to $299\times299$ pixels of 3 channels each.
To autoencode images of this size for the AAE task, we use three different fully convolutional architectures
(Table~\ref{tab:ir2-arch}):

\vskip -0.2in
\begin{itemize}
\item \na0, a small architecture that uses the first few layers of IR2 and loads the
pre-trained parameter values at the start of training the ATN, followed by deconvolutional layers;
\vskip -0.1in
\item \na3, a small architecture that avoids checkerboard artifacts common in deconvolutional
layers by using bilinear resize layers to downsample and upsample between stride 1 convolutions; and
\vskip -0.1in
\item \na2, a medium architecture that is a tower of convolutions followed by deconvolutions.
\end{itemize}
\vskip -0.1in
For the perturbation approach, we use \na0 and \na1, which has many more parameters than the other architectures due to two large fully-connected layers.
The use of fully-connected layers cause the network to learn too slowly for the autoencoding approach (AAE ATN), but can be used to learn perturbations quickly (P-ATN). %in a small number of training steps.

\vskip -0.1in

\paragraph{Hyperparameter search.}
All five architectures across both tasks are trained with the same hyperparameters.
For each architecture and task, we trained four  networks, one for each target class: binoculars, soccer ball, volcano, and zebra.
In total, we trained 20 different ATNs to attack IR2.

To find a good set of hyperparameters for these networks, we did a series of grid searches through reasonable parameter values for learning rate, $\alpha$, and $\beta$, using only Volcano as the target class.
Those training runs were terminated after 0.025 epochs, which is only 1600 training steps with a batch size of 20.
Based on the parameter search, for the results reported here, we set the learning rate to 0.0001, $\alpha=1.5$, and $\beta=0.01$.
All runs were trained for 0.1 epochs (6400 steps) on shuffled training set images, using the Adam optimizer and the TensorFlow default settings. %Slim module~\citep{slim}.

In order to avoid cherrypicking the best results after the networks were trained, we selected four images from the unperturbed validation set to use for the figures in this paper prior to training.
Once training finished, we evaluated the ATNs by passing 1000 images from the validation set through the ATN and measuring IR2's accuracy on those adversarial examples.

\begin{figure*}[p]
  \makebox[\textwidth][c]{
    \includegraphics[width=12.75cm,keepaspectratio]{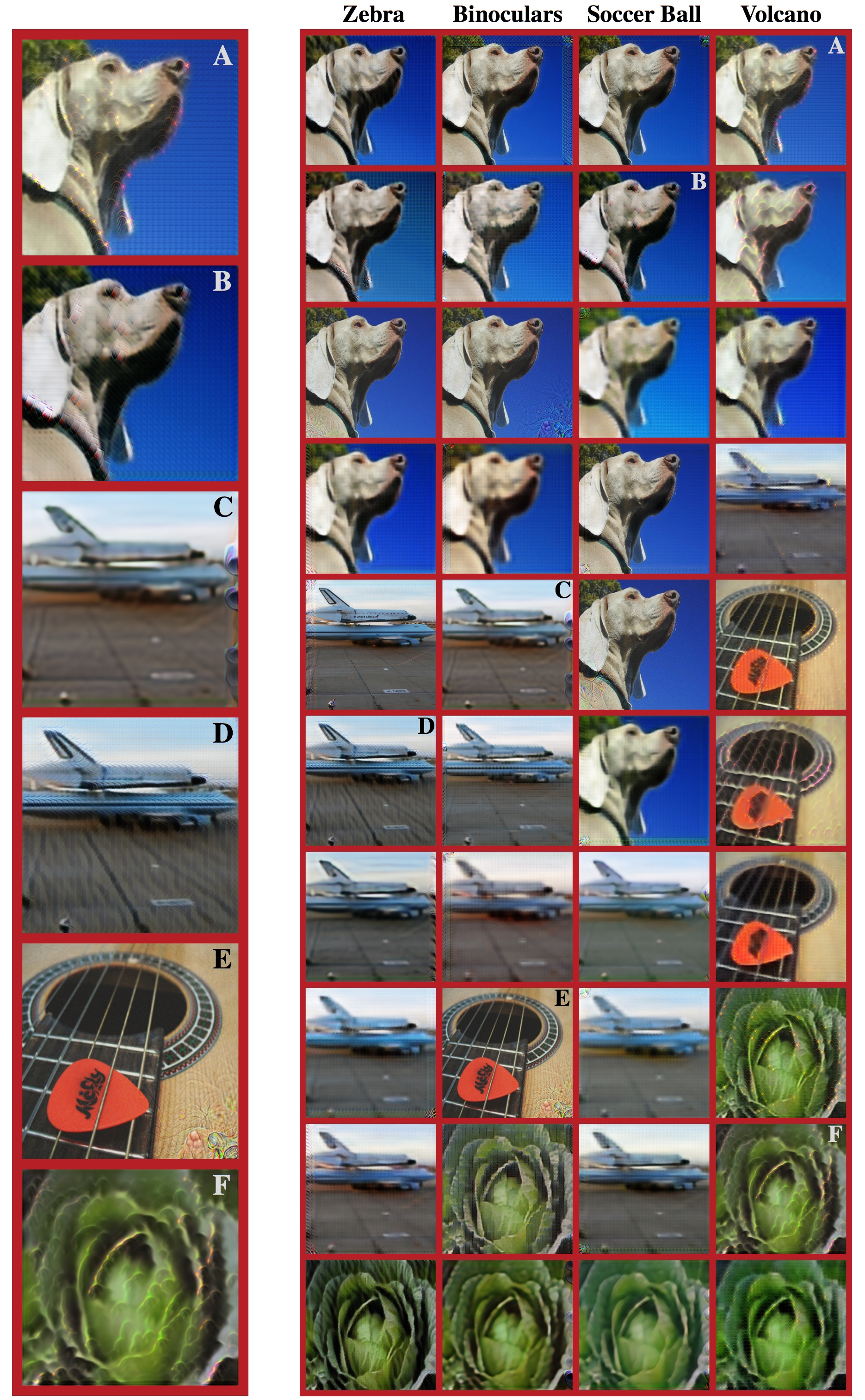}
  }
  \vspace{-0.75cm}
  \caption{\textbf{Adversarial diversity.}
    Left column: selected zoomed samples.
    Right 4 columns: successful adversarial examples for different target classes from a variety of ATNs.
    From the left: Zebra, Binoculars, Soccer Ball, Volcano.
    These images were selected at random from the set of successful adversaries against each target class.
    Unlike existing adversarial techniques, where adversarial examples tend to look alike, these adversarial examples exhibit a great deal of diversity, some of which is quite surprising.
    For example, consider the second image of the space shuttle in the ``Zebra'' column (D).
    In this case, the ATN made the lines on the tarmac darker and more organic, which is somewhat evocative of a zebra's stripes.
    Yet clearly no human would mistake this for an image of a zebra.
    Similarly, the dog's face in (A) has been speckled with a few orange dots (but not the background!), and these are sufficient to convince IR2 that it is a volcano.
    This diversity may be a key to improving the effectiveness of adversarial training, as a more diverse pool of adversarial examples may lead to better network generalization.
    Images A, B, and D are from AAE ATN \na2.
    Images C and F are from AAE ATN \na3.
    Image E is from P-ATN \na1.
    \label{fig:diversity}
  }
\end{figure*}

\subsection{Results Overview}

\begin{figure*}
  \begin{center}

     \includegraphics[width=6.75in,keepaspectratio]{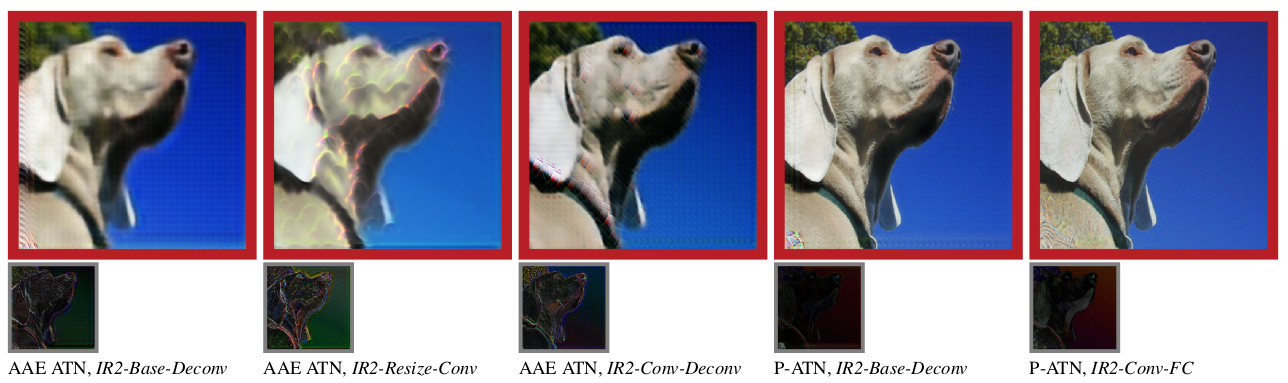}

    \caption{\textbf{Architecture comparisons.}
      Left 3: Adversarial autoencoding ATNs.
      Right 2: Perturbation ATNs.
      All five networks in this figure are trained with the same hyperparameters, apart from the target class, which varies among zebra, soccer ball, and volcano.
      The images in the bottom row show the absolute difference between the original image (not shown) and the adversarial examples.
      There are substantial differences in how the different architectures generate adversarial examples.
      $\na0$ and $\na2$ tend to exhibit checkerboard patterns in their reconstructions, which is a common problem in image generation with deconvolutions in general.
      $\na3$ avoids the checkerboard pattern, but tends to give very smooth outputs.
      $\na0$ performs quite differently when trained as a P-ATN rather than an AAE ATN.
      The P-ATNs focus their perturbations along the edges of the image or in the corners, where they are presumably less likely to disrupt the other top classifications.
      In both P-ATNs, it turns out that the networks learn to mostly ignore the input and simply generate a single perturbation that can be applied to any input image without much change, although the perturbations between networks on the same target image vary substantially.
      This is in stark contrast to the AAE ATNs, which creatively modify each input individually, as seen here and in Figure~\ref{fig:diversity}.
    }
    \label{fig:ir2-arch-comp}
  \end{center}
\end{figure*}

Table~\ref{tab:il2-perf} shows the top-1 adversarial accuracy for each of the 20 model/target combinations.
The AAE approach is superior to the perturbation approach, both in terms of top-1 adversarial accuracy, and in terms of training success.
Nonetheless, the results in Figures~\ref{fig:ir2-arch-comp} and~\ref{fig:diversity} show that using an architecture like $\na1$ can provide a qualitatively different type of adversary from the AAE approach.% which may be superior for particular tasks.
The examples generated using the perturbation approach preserve more pixels in the original image, at the expense of a small region of large perturbations.%  ed with the AAE approach; , so if finding smaller perturbations is a requirement, a P-ATN like $\na1$ may be preferable.

In contrast to the perturbation approaches, the AAE architectures distribute the differences across wider regions of the image.
However, $\na0$ and $\na2$ tend to exhibit checkerboard patterns, which is a common problem in image generation with deconvolutions (\citet{odena2016deconvolution}).
The checkerboarding led us to try $\na3$, which avoids the checkerboard pattern, but gives smooth outputs (Figure~\ref{fig:ir2-arch-comp}).
Interestingly, in all three AAE networks, many of the original high-frequency patterns are replaced with high frequencies that encode the adversarial signal.

The results from $\na0$ show that the same network architectures perform substantially differently when trained as P-ATNs and AAE ATNs.
Since P-ATNs are only learning to perturb the input, these networks are much better at preserving the original image, but the perturbations end up being focused along the edges or in the corners of the image.
The form of the perturbations often manifests itself as ``DeepDream''-like images, as in Figure~\ref{fig:deepdream}.
Approximately the same perturbation, in the same place, is used across all input examples.
Placing the perturbations in that manner is less likely to disrupt the other top classifications, thereby keeping $\LY$ lower.
This is in stark contrast to the AAE ATNs, which creatively modify the input, as seen in Figures~\ref{fig:ir2-arch-comp} and~\ref{fig:diversity}.

\subsection{Detailed Discussion}
\label{sec:ir2-discussion}

\paragraph{Adversarial diversity.}
Figure~\ref{fig:diversity} shows that ATNs are capable of generating a wide variety of adversarial perturbations targeting a single network.
Previous approaches to generating adversarial examples often produced qualitatively uniform results -- they add various amounts of ``noise'' to the image, generally concentrating the noise at pixels with large gradient magnitude for the particular adversarial loss function.
Indeed, \citet{metzen2017} recently showed that it may be possible to train a detector for previous adversarial attacks.
From the perspective of an attacker, then, adversarial examples produced by ATNs may provide a new way past defenses in the cat-and-mouse game of security, since this somewhat unpredictable diversity will likely challenge such approaches to defense.
Perhaps a much more interesting consequence of this diversity is its potential application for more comprehensive adversarial training, as described below.
\vskip -0.25in

\paragraph{Adversarial Training with ATNs.}
In \citet{kurakin2017adversarial}, the authors show the current state-of-the-art in using adveraries for improving training. 
With single step and iterative gradient methods, they find that it is possible to increase a network's robustness to adversarial examples, while suffering a small loss of accuracy on clean inputs.
However, it works only for the adversary the network was trained against.
It appears that ATNs could be used in their adversarial training architecture, and could provide substantially more diversity to the trained model than current adversaries.
This adversarial diversity might improve model test-set generalization and adversarial robustness.

Because ATNs are quick to train relative to the target network (in the case of IR2, hours instead of weeks), reliably produce diverse adversarial examples, and can be automatically checked for quality (by checking their success rate against the target network and the $\LX$ magnitude of the adversarial examples), they could be used as follows:
Train a set of ATNs targeting a random subset of the output classes on a checkpoint of the target network.
Once the ATNs are trained, replace a fraction of each training batch with corresponding adversarial examples, subject to two constraints: the current classifier incorrectly classifies the adversarial example as the target class, and the $\LX$ loss of the adversarial example is below a threshold that indicates it is similar to the original image.
If a given ATN stops producing successful adversarial examples, replace it with a newly trained ATN targeting another randomly selected class.
In this manner, throughout training, the target network would be exposed to a shifting set of diverse adversaries from ATNs that can be trained in a fully-automated manner.\footnote{
  This procedure conceptually resembles GAN training~\citep{goodfellow2014generative} in many ways, but the goal is different: for GANs, the focus is on using an easy-to-train discriminator to learn a hard-to-train generator; for this adversarial training system, the focus is on using easy-to-train generators to learn a hard-to-train multi-class classifier.
}\footnote{%
  Note also that we can run the adversarial example generation in this algorithm on unlabeled data, as described in Section~\ref{arch}.
  \citet{miyato2016distributional} also describe a method for using unlabeled data in a manner conceptually similar to adversarial training.
}

\begin{figure}[t]
  \centering
  \includegraphics[width=3in]{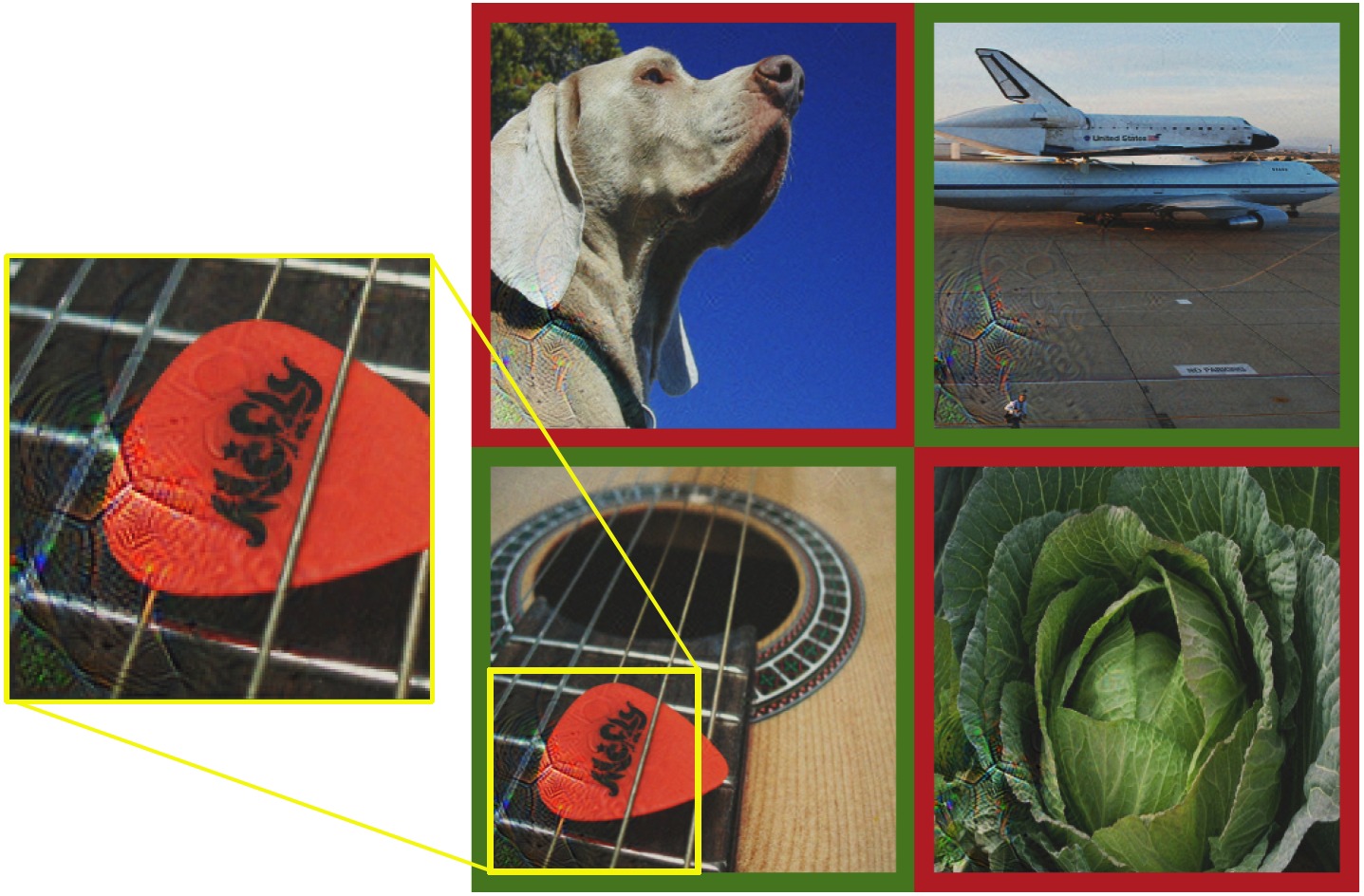}

  \caption{\textbf{DeepDream-style perturbations.}
    Four different images perturbed by $\na1$, targeting soccer ball.
    The images outlined in red were successful adversarial examples against IR2.
    The images outlined in green did not change IR2's top-1 classification.
    The network has learned to add approximately the same perturbation to all images.
    The perturbation resembles part of a soccer ball (lower-left corner).
    The results are akin to those found in DeepDream-like processes~\citep{Mordvinstev2015}.
  }
  \label{fig:deepdream}

  \vspace{\floatsep}

\end{figure}

\paragraph{DeepDream perturbations.}
$\na1$ exhibits interesting behavior not seen in any of the other architectures.
The network builds a perturbation that generally contains spatially coherent, recognizable regions of the target class.
For example, in Figure~\ref{fig:deepdream}, a consistent soccer-ball  ``ghost'' image appears in all of the transformed images.
While the methods and goals of these perturbations are quite different from those generated by DeepDream~\citep{Mordvinstev2015}, the qualitative results appear similar.
$\na1$ seems to learn to distill the target network's representation of the target class in a manner that can be drawn across a large fraction of the image.\footnote{%
  This is likely due to the final fully-connected layer, which has one weight for each pixel and channel, allowing the network to specify a particular output at each pixel.
}
This result hints at a direct relationship between DeepDream-style techniques and adversarial examples that may improve our ability to find and correct weaknesses in our models.

\paragraph{High frequency data.}
The AAE ATNs all remove high frequency data from the images when building their reconstructions.
This is likely to be due to limitations of the underlying architectures.
In particular, all three convolutional architectures have difficulty exactly recreating edges from the input image, due to spatial data loss introduced when downsampling and padding.
Consequently, the $\LX$ loss penalizes high confidence predictions of edge locations, leading the networks to learn to smooth out boundaries in the reconstruction.
This strategy minimizes the overall loss, but it also places a lower bound on the error imposed by pixels in regions with high frequency information.

This lower bound on the loss in some regions provides the network with an interesting strategy when generating an AAE output: it can focus the adversarial perturbations in regions of the input image that have high-frequency noise.
This strategy is visible in many of the more interesting images in Figure~\ref{fig:diversity}.
For example, many of the networks make minimal modification to the sky in the dog image, but add substantial changes around the edges of the dog's face, exactly where the $\LX$ error would be high in a non-adversarial reconstruction.

\begin{table}[t]

  \captionof{table}{%
    ImageNet ATN Architectures.
    \label{tab:ir2-arch}
  }
  \vskip 0.1in

  \makebox[\columnwidth][c]{
  \begin{tiny}

    \setlength{\tabcolsep}{0pt} % Default value: 6pt
    \renewcommand*{\arraystretch}{0.2} % Default value: 1

    \noindent
    \begin{tabular}{lc}

      \hline

      \makecell[l]{$\na0$ \\ (3.4M parameters)} &
      \makecell[r]{
        IR2 MaxPool 5a (35x35x192) $\rightarrow$ Pad (37x37x192) $\rightarrow$ Deconv (4x4x512, stride=2) \\
        $\rightarrow$ Deconv (3x3x256, stride=2) $\rightarrow$ Deconv (4x4x128, stride=2) \\
        $\rightarrow$ Pad (299x299x128) $\rightarrow$ Deconv (4x4x3) $\rightarrow$ Image (299x299x3) \\
      } \\

      \hline

      \makecell[l]{$\na3$ \\ (3.8M parameters)} &
      \makecell[r]{
        Conv (5x5x128) $\rightarrow$ Bilinear Resize (0.5) $\rightarrow$ Conv (4x4x256) $\rightarrow$ \\
        Bilinear Resize (0.5) $\rightarrow$ Conv (3x3x512)
        $\rightarrow$ Bilinear Resize (0.5) $\rightarrow$ Conv (1x1x512) \\
        $\rightarrow$ Bilinear Resize (2) $\rightarrow$ Conv (3x3x256)
        $\rightarrow$ Bilinear Resize (2) $\rightarrow$ Conv (4x4x128) \\
        $\rightarrow$ Pad (299x299x128) $\rightarrow$ Conv (3x3x3) $\rightarrow$ Image (299x299x3) \\
      } \\

      \hline

      \makecell[l]{$\na2$ \\ (12.8M parameters)} &
      \makecell[r]{
        Conv (3x3x256, stride=2) $\rightarrow$ Conv (3x3x512, stride=2) $\rightarrow$ Conv (3x3x768, stride=2) \\
        $\rightarrow$ Deconv (4x4x512, stride=2) $\rightarrow$ Deconv (3x3x256, stride=2) \\
        $\rightarrow$ Deconv (4x4x128, stride=2) $\rightarrow$ Pad (299x299x128) \\
        $\rightarrow$ Deconv (4x4x3) $\rightarrow$ Image (299x299x3) \\
      } \\

      \hline

      \makecell[l]{$\na1$ \\ (233.7M parameters)} &
      \makecell[r]{
        Conv (3x3x512, stride=2) $\rightarrow$ Conv (3x3x256, stride=2) $\rightarrow$ Conv (3x3x128, stride=2) \\
        $\rightarrow$ FC (512) $\rightarrow$ FC (268203) $\rightarrow$ Image (299x299x3) \\
      } \\

      \hline

    \end{tabular}

  \end{tiny}
  } % end \makebox[\columnwidth][c]

  \vskip -0.1in
  \vspace{\floatsep}
  \vskip -0.1in

  \begin{center}
  \begin{tiny}
    \captionof{table}{%
      IL2 ATN Performance
      \label{tab:il2-perf}
    }
    \vskip 0.1in
  \begin{sc}
    \begin{tabular}{l|cccc}

      \hline
      & \multicolumn{4}{c}{\scriptsize{P-ATN Target Class Top-1 Accuracy}} \\
      & Binoculars & Soccer Ball & Volcano & Zebra \\

      \hline

      \scriptsize{\na0} &
      \makecell{\scriptsize{        66.0\%} } &
      \makecell{\scriptsize{        56.5\%} } &
      \makecell{\scriptsize{         0.2\%} } &
      \makecell{\scriptsize{        43.2\%} } \\

      \scriptsize{\na1} &
      \makecell{\scriptsize{\textbf{79.9\%}}} &
      \makecell{\scriptsize{\textbf{78.8\%}}} &
      \makecell{\scriptsize{         0.0\%} } &
      \makecell{\scriptsize{\textbf{85.6\%}}} \\

      \hline
      \hline

      & \multicolumn{4}{c}{\scriptsize{AAE ATN Target Class Top-1 Accuracy}} \\
      & Binoculars & Soccer Ball & Volcano & Zebra \\

      \hline

      \scriptsize{\na0} &
      \makecell{\scriptsize{\textbf{83.0\%}}} &
      \makecell{\scriptsize{\textbf{92.1\%}}} &
      \makecell{\scriptsize{        88.1\%} } &
      \makecell{\scriptsize{\textbf{88.2\%}}} \\

      \scriptsize{\na3} &
      \makecell{\scriptsize{        69.8\%} } &
      \makecell{\scriptsize{        61.4\%} } &
      \makecell{\scriptsize{\textbf{91.1\%}}} &
      \makecell{\scriptsize{        80.2\%} } \\

      \scriptsize{\na2} &
      \makecell{\scriptsize{        56.6\%} } &
      \makecell{\scriptsize{        75.0\%} } &
      \makecell{\scriptsize{        87.3\%} } &
      \makecell{\scriptsize{        79.1\%} } \\

      \hline

    \end{tabular}
  \end{sc}
  \end{tiny}
  \end{center}
  \vskip -0.1in

\end{table}

